\documentclass[sigconf]{acmart}

\usepackage{multirow}
\usepackage{hhline}

\AtBeginDocument{%
  \providecommand\BibTeX{{%
    \normalfont B\kern-0.5em{\scshape i\kern-0.25em b}\kern-0.8em\TeX}}}

\copyrightyear{2020}
\acmYear{2020}
\setcopyright{acmcopyright}\acmConference[KDD '20]{Proceedings of the 26th ACM SIGKDD Conference on Knowledge Discovery and Data Mining}{August 23--27, 2020}{Virtual Event, CA, USA}
\acmBooktitle{Proceedings of the 26th ACM SIGKDD Conference on Knowledge Discovery and Data Mining (KDD '20), August 23--27, 2020, Virtual Event, CA, USA}
\acmPrice{15.00}
\acmDOI{10.1145/3394486.3403110}
\acmISBN{978-1-4503-7998-4/20/08}

\begin{document}
\fancyhead{}




\title{Deep Learning of High-Order Interactions for Protein Interface Prediction}


\author{Yi Liu}
\affiliation{%
  \institution{Texas A\&M University}
  \city{College Station}
  \state{TX}}
\email{yiliu@tamu.edu}

\author{Hao Yuan}
\affiliation{%
  \institution{Texas A\&M University}
  \city{College Station}
  \state{TX}}
\email{hao.yuan@tamu.edu}

\author{Lei Cai}
\affiliation{%
  \institution{Washington State University}
  \city{Pullman}
  \state{WA}}
\email{lei.cai@wsu.edu}

\author{Shuiwang Ji}
\affiliation{%
  \institution{Texas A\&M University}
  \city{College Station}
  \state{TX}}
\email{sji@tamu.edu}

\renewcommand{\shortauthors}{Liu, et al.}

\begin{abstract}
Protein interactions are important in a broad range of biological
processes. Traditionally, computational methods have been developed
to automatically predict protein interface from hand-crafted
features. Recent approaches employ deep neural networks and predict
the interaction of each amino acid pair independently. However,
these methods do not incorporate the important sequential
information from amino acid chains and the high-order pairwise
interactions. Intuitively, the prediction of an amino acid pair
should depend on both their features and the information of other
amino acid pairs. In this work, we propose to formulate the protein
interface prediction as a 2D dense prediction problem. In addition,
we propose a novel deep model to incorporate the sequential
information and high-order pairwise interactions to perform
interface predictions. We represent proteins as graphs and employ
graph neural networks to learn node features. Then we propose the
sequential modeling method to incorporate the sequential information
and reorder the feature matrix. Next, we incorporate high-order
pairwise interactions to generate a 3D tensor containing different
pairwise interactions. Finally, we employ convolutional neural
networks to perform 2D dense predictions. Experimental results on
multiple benchmarks demonstrate that our proposed method can
consistently improve the protein interface prediction performance.
\end{abstract}




\maketitle

\section{Introduction}
Protein interactions play an important role in biological processes.
Interacted proteins form complicated protein networks known as
protein complexes, which can perform a vast range of biological
functions~\cite{zhou2007interaction}. Protein interactions occur via
interfaces, bonds of amino acids from different proteins. Locating
protein interfaces requires to identify all amino acid pairs, which
is an important yet challenging
problem~\cite{jordan2012predicting,northey2018intpred}. Experimental
identification is expensive and
time-consuming~\cite{ezkurdia2009progress}. Computational
methods~\cite{ahmad2011partner,liu2009prediction,vsikic2009prediction,wu2006comparing,bradford2005improved,sanchez2019bipspi}
have been proposed to automatically predict protein interfaces.
These methods focus on constructing hand-crafted features from
different domains, then applying conventional machine learning
approaches for interface prediction.

Deep learning methods have shown great success on grid-like data
such as texts~\cite{kvaswani2017attention,liu2019learning},
images~\cite{ronneberger2015u,simonyan2014very,liu2020global}, and non-grid data
such as graphs~\cite{hamilton2017inductive,kipf2016semi,yuan2020structpool}. Following
the success, recent studies~\cite{fout2017protein,townshend2019end}
propose to apply deep learning methods to learn features for amino
acids and perform interaction predictions. Existing
work~\cite{townshend2019end} folds proteins into 4D grid-like data
and employ 3D Convolutional Neural Networks (CNNs) for feature
learning. This is followed by dense layers to determine whether the
two amino acids interact with each other. However, the topological
structure information is ignored by representing proteins as
grid-like data. Such information is important to decide the inherent
properties of amino acids and proteins. In addition, recent
work~\cite{fout2017protein} represents proteins as graphs, where
nodes are amino acids and edges are affinities between nodes. Then
it applies Graph Neural Networks (GNNs) to learn node features. For
any amino acid pair, the node features are concatenated and a
classifier is built based on these features. However, the original
sequential information from amino acid chains is ignored when
converting from proteins to graphs. In addition, existing studies
predict each amino acid pair separately such that only the
information from the input amino acid pair is considered. Due to the
complex structure of proteins, the prediction of a amino acid pair
may also depend on other amino acids. Such relations in known as the
high-order pairwise interactions while none of existing work
explicitly incorporate them.

To overcome these limitations, we propose a novel framework to solve
the protein interface prediction problem. Instead of identifying
each amino acid pair independently, we formulate it as a 2D dense
prediction problem, which predicts all possible pairs
simultaneously. In addition, we propose a novel deep learning model
to solve it. Similar to existing work~\cite{fout2017protein}, we
also represent proteins as graphs and employ GNNs to aggregate
neighborhood information and learn node features. To incorporate the
sequential information of amino acid chains,  we propose the
sequential modeling to reorder the features and preserve original
sequential information. Such a step also enables the use of
convolution operations in the latter stage. Next, we construct the
high-order pairwise feature interactions based on node features,
resulting in a 3D tensor. For each location, it stores the feature
interaction of the corresponding amino acid pair. Note that the
sequential information is also preserved in this tensor. Then 2D
CNNs are employed to extract high-level pairwise interaction
patterns and make predictions for all pairs as a 2D dense prediction
task. Furthermore, to address the data imbalance problem, we not
only incorporate cross-protein amino acid pairs for training, but
also involve in-protein amino acid pairs. We evaluate our methods on
the three cross-protein docking and binding affinity
benchmarks~\cite{hwang2010protein,vreven2015updates,hwang2008protein}.
Experimental results show that our methods consistently outperform
all state-of-the-art methods and achieve over $3\%$ performance
improvement. The results demonstrate the effectiveness of our
proposed sequential modeling and high-order pairwise interaction
method that incorporate both sequential information and high-order
pairwise interaction patterns. Overall, our major contributions are
summarized as follows:
\begin{itemize}
    \item We propose a novel formulation for the protein interface prediction problem. We consider it as a 2D dense prediction problem,
    which is a structured prediction problem and predicts all amino acid pair simultaneously.
    \item We propose a novel deep learning method, which captures structural information from protein graphs, sequential information from original
    amino acid chains, and high-order pairwise interaction information between different amino acid pairs.
    \item We obtain the new state-of-the-art performance on three protein docking and binding affinity benchmarks. Experimental results show the effectiveness of our proposed sequential modeling and high-order pairwise interaction method.
\end{itemize}

\section{Related Work}

Protein interface prediction has been studied intensively.
To predict protein interfaces between protein complexes, two categories of methods have been proposed, those are, partner-independent prediction
and partner-specific prediction~\cite{afsar2014pairpred}. The former is to
predict whether there is an interaction between an amino acid in the given protein with any other protein~\cite{leis2010silico,deng2009prediction,zhou2007interaction}.
The latter is to predict if there is an interaction between any two amino acids
from two different proteins~\cite{sanchez2019bipspi,northey2018intpred}.
Partner-specific prediction has been demonstrated to achieve better performance
due to the use of interaction information in protein complexes~\cite{afsar2014pairpred,ahmad2011partner}.
In this work, we focus on the partner-specific prediction
that we predict interactions between any two amino acids from different proteins.

There exist several families of techniques for partner-specific prediction.
Template-based methods predict interfaces in a query complex by computing interface and structure
similarities to a given template complex~\cite{tuncbag2011prediction}.  One limitation is that the prediction can be performed only when there exit template complexes.
Docking methods~\cite{pierce2011accelerating,schueler2005progress} typically predict several possible protein complexes from the given proteins,
then use ranking criteria to decide the most native one. The interface is then
identified after deciding the specific complex structure. Docking methods have shown
similar performance with the template-based methods. Another big family is the machine learning-based
methods. These methods focus on constructing features then use machine learning techniques for
classification. The features are sourced from different domains. The used machine learning
techniques include SVM~\cite{bradford2005improved} and
random forests~\cite{segura2012holistic}, etc.

The latest developed methods~\cite{fout2017protein,townshend2019end} emphasize the
representations of proteins and amino acids and have achieved the best reported performance.
In the work~\cite{townshend2019end}, an amino acid is represented as 4D grid-like data.
The first 3 dimensions are the spatial coordinates of all atoms in the amino acid and
the last one indicates the types of atoms. Then 3D CNNs are employed for amino acid feature learning
and dense layers are to predict if the two amino acids interact or not.
The work~\cite{fout2017protein} represents proteins as graphs and use GNNs to aggregate structural information,
followed by the dense layers for binary classification. However, using dense layers to classify each
amino acid pair independently neglects
the important high-order pairwise interactions that the prediction of one amino acid pairs may depend on
the information from other amino acid pairs.
In addition, the former fails to consider topological structure information by representing proteins
as grid-like data, whereas the latter does not incorporate sequential information by representing proteins as
graphs which are order-invariant.

Most methods for protein interface prediction are evaluated on the family of
Docking Benchmark (DB) datasets, which contain DB2~\cite{mintseris2005protein}, DB3~\cite{hwang2008protein}, DB4~\cite{hwang2010protein} and DB5~\cite{vreven2015updates}.
The first three are all the subsets of DB5, which is the largest and most recent benchmark
dataset for interface prediction. It also contains the most labeled examples for the problem.
There are 230 protein complexes in total, and the the number of labeled interacted amino acid pairs
is 20875. All the proteins carry structural and sequential information. Currently, DB5
is the most popular dataset for protein interface prediction, like the ImageNet~\cite{ILSVRC15} in the computer vision domain.
Before DB5 was generated, DB4 and DB3 were intensively used to evaluate different methods.

To overcome limitations in existing works,
we propose an end-to-end framework that incorporates structural and sequential information
and high-order pairwise interaction patterns for protein interface prediction.
We conduct experiments on three Docking Benchmark datasets to demonstrate the effectiveness of our proposed methods.

\section{Methods}
The  structural and sequential information are both
important to determine the properties of proteins.
However, existing work~\cite{fout2017protein} represents proteins as graphs
which can only capture structural information but neglect the sequential information of the original amino acid chains.
In addition, existing methods~\cite{fout2017protein,townshend2019end,sanchez2019bipspi} classify each amino acid pair
separately, through which only the information of input amino acid pair is considered.
However, the prediction of an input amino acid pair may also depend on the information of other amino acid pairs. Such relations are known as
high-order pairwise interactions, and none of existing work explicitly considers them.
To incorporate all of structural information, sequential information, and high-order pairwise interactions,
we propose a novel formulation for the protein interface prediction problem and a novel deep learning method to solve it.

\subsection{Problem Formulation}
A protein complex is composed of two proteins, known as the ligand protein and the receptor protein.
Suppose the ligand protein has $n_l$ amino acids and
the receptor protein has $n_r$ amino acids, there are $n_l\times n_r$
possible amino acids pairs.
Protein interface prediction problem aims at predicting
if there exists an interaction within each amino acid pair.
Following the existing work~\cite{fout2017protein}, we represent proteins as graphs,
where nodes represent amino acids and edges indicate affinities
between amino acids.
Formally, we define the feature matrix of the ligand protein as
$\hat{H}^l = [\hat{\mathbf{h}}^l_{1}, \hat{\mathbf{h}}^l_{2}, \cdots, \hat{\mathbf{h}}^l_{n_l}]\in\mathbb{R}^{d\times n_l}$
and the feature matrix of the receptor protein as
$\hat{H}^r = [\hat{\mathbf{h}}^r_{1}, \hat{\mathbf{h}}^r_{2}, \cdots, \hat{\mathbf{h}}^r_{n_r}]\in\mathbb{R}^{d\times n_r}$
where $d$ denotes that each node has a $d$-dimensional feature vector.

The existing work~\cite{fout2017protein} formulates it as a binary classification problem
that predicts the interaction between each node pair separately. Specifically, for the
the $i$-th node in $\hat{H}^l$ and the $j$-th node in $\hat{H}^r$, it
concatenates the corresponding feature vectors $\hat{\mathbf{h}}^l_{i}$ and $\hat{\mathbf{h}}^r_{j}$
, then uses dense layers as a binary classifier to determine whether the
two nodes interact with each other. However, such a formulation ignores high-order pairwise interactions that
the interaction prediction of an amino acid pair may also depend on the information of other amino acid pairs.
In addition, converting amino acids to graphs loses the sequential information of the original amino acid chains.

To address these issues, we incorporate the sequential information and formulate the protein interface prediction as a 2D dense prediction problem. First, given the node feature matrices $\hat{H}^l$ and $\hat{H}^r$,
we propose the sequential modeling (SM) to restore the sequential information, which results
in the order-preserved node feature matrix
$H^l = [\mathbf{h}^l_{1}, \mathbf{h}^l_{2}, \cdots, \mathbf{h}^l_{n_l}]\in\mathbb{R}^{d\times n_l}$ for the ligand protein and
$H^r = [\mathbf{h}^r_{1}, \mathbf{h}^r_{2}, \cdots, \mathbf{h}^r_{n_r}]\in\mathbb{R}^{d\times n_r}$ for the receptor protein.
Next, we propose the high-order pairwise interaction (HOPI) to generate a 3D tensor $\mathcal{Q}\in\mathbb{R}^{n_l\times n_r\times c}$ where each $\mathcal{Q}_{ij}\in\mathbb{R}^{c}$ denotes the feature combination of
the $i$-th node in $H^l$ and the $j$-th node in $H^r$.
Finally, based on the tensor $Q$, the protein interface prediction problem predicts a 2D matrix
$O\in \{0,1\}^{n_l\times n_r}$. Each element of $O$ can be either 0 or 1.
For location $O_{i,j}$, 1 means there exists an interaction between $i$-th amino acid of the ligand protein
and $j$-th amino acid of the receptor protein, while 0 indicates
there is no interaction. Since the predictions $O$ is generated based on the whole tensor $Q$, both sequential information and
high-order pairwise interactions are incorporated.

\begin{figure}[t] \includegraphics[width=\columnwidth]{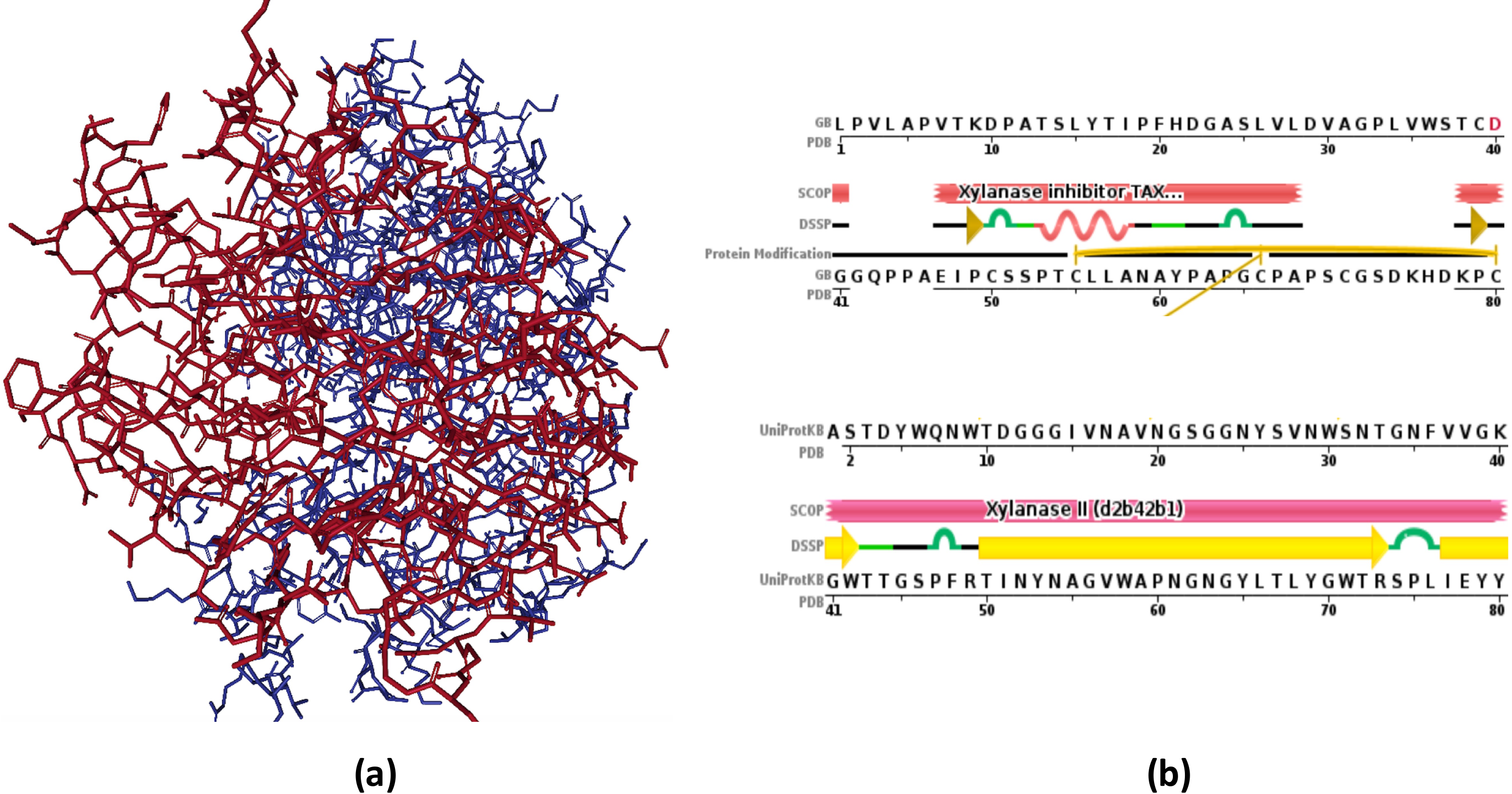}
\caption{Structure view and sequence view of the protein complex
2B42. The protein complex contains two proteins, know as the ligand protein and the receptor protein.
(a) is the structure view, where the red one denotes
the ligand protein and the blue one denotes the receptor protein.
(b) shows the first 80 amino acids in the amino acid sequence chains
of the ligand protein and the receptor protein at the top and bottom, respectively.
The figures are sourced from the Protein Data Bank website
\href{https://www.rcsb.org/}{https://www.rcsb.org/} and~\cite{pollet2009identification,drew1981structure}.
}
\label{fg:strseq}
\end{figure}

\begin{figure*}[t] \includegraphics[width=0.9 \textwidth]{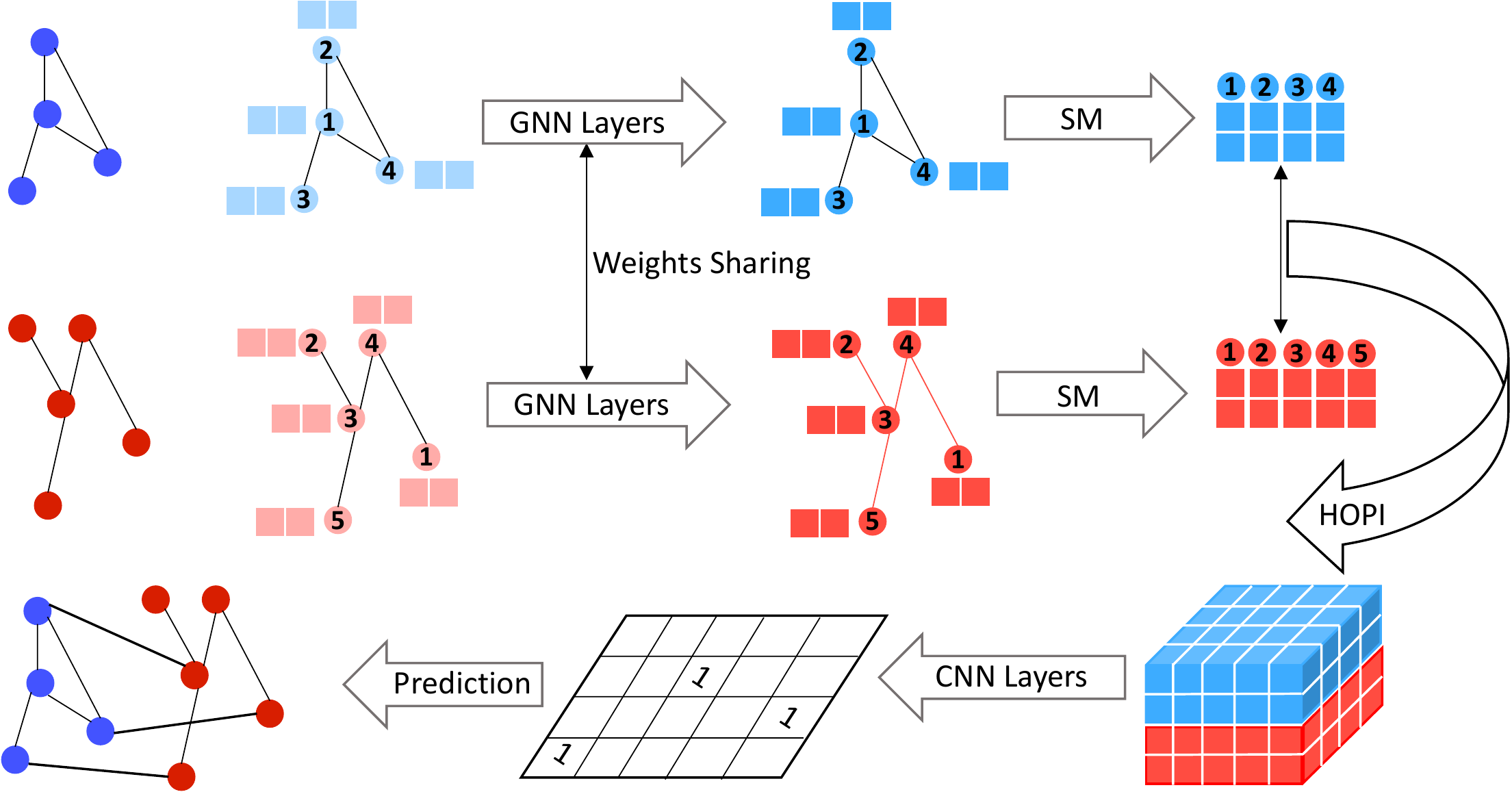}
\caption{The overall architecture of our proposed methods. Given a ligand protein and
a receptor protein, the task is to predict the interface between them. The two proteins are
represented as graphs, where nodes represent amino acids and edges indicate affinities between nodes.
Here we use the example that one graph contains 4 nodes and the other contains 5, and each node has 2 features. GNN layers are used to aggregate structural
information and the weighs are shared by the two graphs. After that, sequential modeling (SM) is used to restore the
sequential information from the original amino acid chains, resulting in two node feature matrices with the dimensions of
$2 \times 4$ and $2 \times 5$, respectively. Then the high-order pairwise interaction (HOPI) is performed on these two matrices
to build pairwise interactions for any two amino acids from different proteins. Concatenation is used to build
interactions between two amino acids.
The achieved 3D tensor is with the dimension of $4 \times 5 \times 4$ and stores in-protein structural and sequential information and cross-protein
pairwise interactions. Then 2D CNN layers are used for dense prediction, which produces the output feature map with the dimension of $4 \times 5$. The pixel
value of each pixel on the output is either 1 or 0 to indicate an interaction for the corresponding amino acid pair.
}
\label{fg:arch}
\end{figure*}

\subsection{Sequential Modeling}
Both structural and sequential information are important
for studying properties of protein complexes. As shown in Figure~\ref{fg:strseq},
representing proteins as graphs can well-convey the structural information. It is a popular way since
we can employ graph neural networks to pass, transform and aggregate structural information across
graphs~\cite{ying2018hierarchical}.
However, the sequential information from the original amino acid chains is lost when converting proteins to graphs because of
order-invariant property of graphs.
Such a sequential structure is
the primary structure of a protein, which is unique to other proteins and defines important functions of the protein.

To overcome this limitation, we propose the sequential modeling to preserve the sequential information of the original
input amino acid chains for a given protein. Formally, given an input protein with $m$ amino acids, we first record the original sequential order set $I = (1,2,\cdots,m) $. Then we formulate it as graphs and map each node in the graph with the index in $I$. Next,
we employ graph neural networks to learn node features, denoted as $\hat{H} = [\hat{\mathbf{h}}_{1}, \hat{\mathbf{h}}_{2}, \cdots, \hat{\mathbf{h}}_{m}]\in\mathbb{R}^{d\times m}$, where $d$ is the dimension of a node feature vector.
Then we reorder the feature matrix $\hat{H}$ based on the order set $I$ as
\begin{equation}\label{eq:reorder}
H = \mbox{reorder}(\hat{H}, I)\in\mathbb{R}^{d\times m}.
\end{equation}
Then the node feature vectors in the new feature matrix $H$ has a consistent order with the original amino acid
sequential order.
In this way, feature matrix $H$ successfully captures both structural information
from the protein graph and the sequential information from the amino acid chains.
We believe such a reordering operation helps capture complex inherent relationships among amino acids,
thereby resulting in more accurate interface prediction.

\subsection{High-Order Pairwise Interactions} \label{sec:HOPI}
Protein interface prediction aims at determining interactions between two amino acids from different proteins.
The protein structure is the 3D arrangements in amino acid chains and always folds into specific spatial conformations to enable
biological functions~\cite{pauling1951structure}. It is possible that any two amino acids from different proteins can interact with each other.
Existing methods~\cite{fout2017protein,sanchez2019bipspi,townshend2019end} predict the interaction
for each amino acid pair separately.
One amino acid is picked from the ligand protein graph and the other is
from the receptor protein graph. The features of the two amino acids are concatenated and passed
to dense layers for binary classification.
However, high-order context for amino acid pairs are ignored, which can help extract the important high-level interaction patterns.
Hence, we propose the high-order pairwise interactions to learn complex interaction patterns
for interface prediction.

Suppose we have the sequence-preserved node feature matrix
$H^l = [\mathbf{h}^l_{1}, \mathbf{h}^l_{2}, \cdots, \mathbf{h}^l_{n_l}]\in\mathbb{R}^{d\times n_l}$ for the ligand protein and
$H^r = [\mathbf{h}^r_{1}, \mathbf{h}^r_{2}, \cdots, \mathbf{h}^r_{n_r}]\in\mathbb{R}^{d\times n_r}$ for the receptor protein.
We compute a third-order tensor $\mathcal{Q}\in\mathbb{R}^{n_l\times n_r\times c}$. Each $\mathcal{Q}_{ij}\in\mathbb{R}^{c}$ in $\mathcal{Q}$ is the transformation of $\mathbf{h}^l_{i}$ and $\mathbf{h}^r_{j}$.
It can be computed by either summation of $\mathbf{h}^l_i$ and $\mathbf{h}^r_j$ or the concatenation of $\mathbf{h}^l_i$ and $\mathbf{h}^r_j$.
The proposed HOPI allows the tensor $\mathcal{Q}$ to store structural information, sequential information, and inherent high-order pairwise interactions information. Then we employ convolutional neural networks (CNNs) to perform 2D dense predictions based on the tensor $\mathcal{Q}$.
Stacking several CNN layers
extracts high-level pairwise interaction patterns from a region containing a
subsequence from the ligand protein, a
subsequence from the receptor protein, and inherent high-order pairwise interactions from the two subsequences.
Finally, the output $O\in\{0,1\}^{n_l\times n_r}$ indicates the interactions between any possible amino acid pairs.
Note that the prediction of $O_{i,j}$ depends not only on  $\mathcal{Q}_{ij}$ but also on all feature interactions within its receptive field.

The overall architecture is illustrated in Figure~\ref{fg:arch}.
Given a ligand protein and a receptor protein,
GNNs are used to pass, transform and aggregate the structural information in protein graphs.
All GNN layers are shared by the two proteins.
Then the proposed sequential modeling performs reordering to preserves the sequential information of the original amino acid chains for both proteins.
Next, the proposed high-order pairwise interaction method produce a 3D tensor containing feature interactions for all amino acid pairs.
Finally, the tensor is  passed to 2D CNNs for 2D dense prediction. The output map
contains interaction predictions for all amino acid pairs.
Note that any modern CNN architecture such as ResNet~\cite{he2016deep}, UNet~\cite{ronneberger2015u} or DeepLab~\cite{chen2017deeplab} can be flexibly integrated into our framework to perform dense prediction and the whole system can be trained end-to-end.

\subsection{Graph Neural Networks}
We employ graph neural networks to aggregate structural information.
Suppose a node $\mathbf{h}_i$ in the protein graph has $n$
nodes in its neighborhood. The neighboring node feature matrix is $H_i\in\mathbb{R}^{d^N\times n}$, and the
neighboring edge feature matrix is $E_i\in\mathbb{R}^{d^E\times n}$, where $d^N$ is the dimension of node feature vectors and $d^E$ is the dimension of edge feature vectors. We first aggregate both
node and edge features from neighborhood as

\begin{equation}\label{eq:node_edge_feature}
Z_i = \mbox{tanh}(W^NH_i+W^EE_i) \in\mathbb{R}^{d^N\times n},
\end{equation}
where $W^N\in\mathbb{R}^{d^N\times d^N}$, $W^E\in\mathbb{R}^{d^N\times d^E}$, and
$\mbox{tanh}(\cdot)$ is an element-wise operation.
$W^N$ and $W^E$ are used to perform linear transformation on the neighboring node features
and edge features, respectively.
The node feature matrix $H_i$ and edge feature matrix $E_i$ can be treated as
a set of node vectors $[\mathbf{h}_{i1}, \cdots, \mathbf{h}_{ij}, \cdots, \mathbf{h}_{in}]$ and a set of edge vectors
$[\mathbf{e}_{i1}, \cdots, \mathbf{e}_{ij}, \cdots, \mathbf{e}_{in}]$, respectively.
Note that node vectors in $H_i$ and edge vectors in $E_i$ are in a consistent order. An edge
$\mathbf{e}_{ij}$ links the center node $\mathbf{h}_{i}$ to the corresponding node $\mathbf{h}_{ij}$.
The achieved matrix $Z_i$ aggregates information from both nodes and edges within the neighborhood.
Each vector $\mathbf{z}_{ij}\in\mathbb{R}^{d^N}$ contains information from the node $\mathbf{h}_{ij}$
and the edge $\mathbf{e}_{ij}$.

We use two methods to transform neighborhood information to the center node. The first one is to simply perform average on
all neighboring vectors $\mathbf{z}_{ij}, j = 1,\cdots, n$, namely neighborhood average (NeiA). Another approach is neighborhood weighted average (NeiWA), which essentially assigns relatively larger weights to these important
vectors while smaller weights to the ones that are less important, then performs the weighted average on the neighboring nodes and edges. We introduce the two approaches below.
\subsubsection{Neighborhood Average}
For a node $\mathbf{h}_i$ in the protein graph, the output of a GNN layer with NeiA is computed as
\begin{equation}\label{eq:residul_weisum}
\hat{\mathbf{h}}_i = \mathbf{h}_i + \frac{1}{n}Z_i\mathbf{1}_n\in\mathbb{R}^{d^N},
\end{equation}
where $\mathbf{1}_n\in\mathbb{R}^{n}$
denotes a vector of all ones of dimension $n$. Essentially,
we perform average across all the vectors in $Z_i$, and the final output
is obtained by adding it with the residual identity map of the input.
\subsubsection{Neighborhood Weighted Average}
It is natural to consider that not all entries in $Z_i$ contribute equally when aggregating neighboring information.
We want to grant larger weights to these important node and edges for the center node. Formally, for a given node $\mathbf{h}_i$, the node-wise forward propagation of a
GNN layer with NeiWA is computed as
\begin{equation}\label{eq:att_weight}
\mathbf{a} = \mbox{softmax}(Z^T_i\mathbf{q})\in\mathbb{R}^{n},
\end{equation}

\begin{equation}\label{eq:residul_weisum}
\hat{\mathbf{h}}_i = \mathbf{h}_i + \frac{1}{n}Z_i\mathbf{a}\in\mathbb{R}^{d^N},
\end{equation}
where
$\mathbf{q}\in\mathbb{R}^{d^N}$ is a trainable vector which is trained during the whole training process,
and $\mbox{softmax}(\cdot)$ is an element-wise softmax operation.
Basically, projection from each vector in $Z_i$ to the
trainable vector $\mathbf{q}$ is performed to compute the weight vector $\mathbf{a}$,
in which each entry is the importance score for the corresponding vector in $Z_i$. After this,
the weighted average is performed on $Z_i$ to aggregate information from more informative nodes and edges to the center node. The final output is achieved by adding back the input node features.

\subsection{Incorporating In-Protein Pairwise Interactions}

\begin{figure}[t] \includegraphics[width=0.6\columnwidth]{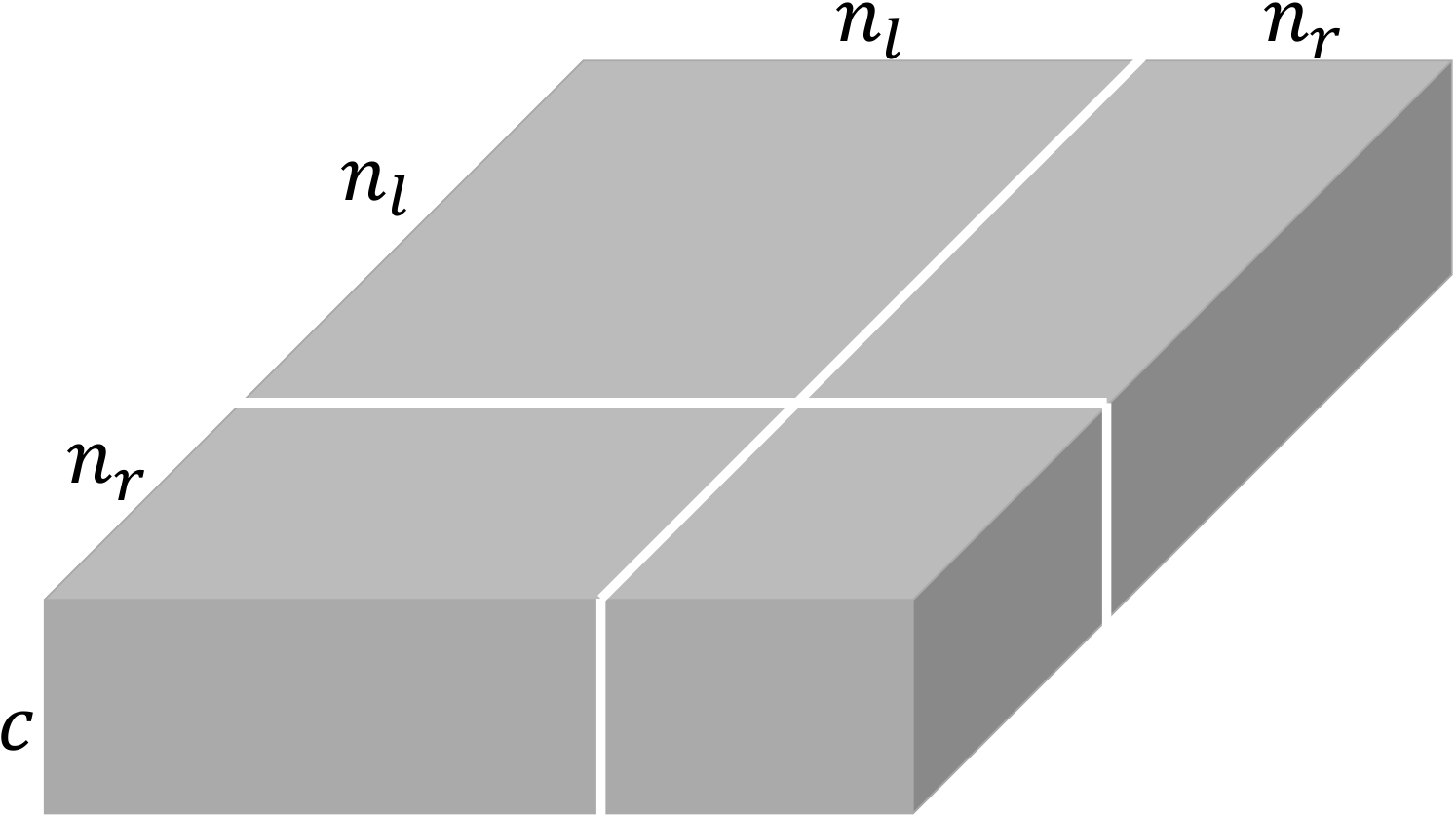}
\caption{An illustration of incorporating in-protein pairwise information
to the tensor $\mathcal{Q}$. $n_l$ denotes the number of nodes
in the ligand protein and $n_r$ denotes the number of nodes
in the receptor protein. $c$ is the number of channels. The
$n_l \times n_l$ patch stores in-protein structural and sequential information
and in-protein pairwise interactions in the ligand protein. The $n_r \times n_r$ patch is similar for the receptor protein. Two $n_l \times n_r$ patches
store in-protein structural and sequential information
and cross-protein pairwise interactions.}
\label{fg:inpro}
\end{figure}

Protein interface prediction is to determine whether there are interactions
between amino acids from two different proteins. Essentially, it
investigates cross-protein pairwise interactions.
The interactions can be partly
determined by features and inherent properties of amino acids in both proteins.
We name an interacted amino acid pair as a positive sample and a non-interacted pair as a negative sample.
Generally, the number of positive samples is significant less than that of the negative samples
in a protein complex. Hence, it causes the data unbalance problem. To address this issue, we propose to
incorporate in-protein interaction information to increase the number of positive examples, and hence improve the predictions of cross-protein interactions.

Specifically, we propose to use HOPI to capture both in-protein
and cross-protein pairwise interactions.
Given the sequence-reserved node feature matrix
$H^l\in\mathbb{R}^{d\times n_l}$ for the ligand protein and
$H^r\in\mathbb{R}^{d\times n_r}$ for the receptor protein, the achieved tensor $\mathcal{Q}$ is expanded to the size of
$(n_l+n_r)\times (n_l+n_r)\times c$. An illustration of the tensor $\mathcal{Q}$ is provided
in Figure~\ref{fg:inpro}.
Either of the two regions $n_l\times n_l$ and $n_r\times n_r$
contains in-protein structural and sequential information,
and in-protein pairwise interactions.
The two regions $n_l\times n_r$ are same as those introduced in Section~\ref{sec:HOPI}, which contain cross-protein interactions.
2D CNNs are performed such that in-protein structural and sequential information, in-protein pairwise interactions, and cross-protein pairwise interactions are all captured for interface prediction.

\subsection{Training Strategies}
We define a training sample as a pair of amino acids and is labeled by their interaction.
All samples in a protein complex can be treated as a sub-epoch.
During one iteration, the network is trained on a part of samples in a protein complex.
In this way, two subgraphs on the ligand and the receptor graphs are updated when aggregating neighboring structural information using GNNs. A small patch is updated on the tensor
$\mathcal{Q}$ when performing dense prediction using CNNs.
A sub-epoch is finished when all training samples in one graph complex are used. And
an epoch is finished when all training samples in all graph complexes in the dataset are
passed to the network.

When considering the cross-protein pairwise interactions
only, a common situation is that the number of positive samples
is much less than that of the negative samples in a protein complex.
Down-sampling on the negative samples is usually
required to reach a reasonable predefined positive-negative (PN) ratio, which
allows the training of the network. However, all negative samples are kept in the
prediction phase.
One way for data augmentation is to incorporate
in-protein pairwise interactions. By doing this,
some in-protein positive samples could be added
in the training process. This is expected to compensate
the small PN ratio in
the cross-protein pairwise interactions.

\section{Experimental Studies}
\subsection{Datasets}
We use three datasets to evaluate our proposed methods.
They all come from the popular Docking Benchmarks,
which include several protein-protein docking and binding affinity benchmarks.
The first one is generated from Docking Benchmarks version 5 (DB5)~\cite{vreven2015updates,fout2017protein}.
It is the largest and the most recent dataset which contains
complicated protein structures for protein interface prediction.
The dataset was originally split into training, validation and test sets~\cite{fout2017protein,townshend2019end,sanchez2019bipspi}.
The training set contains 140 complexes, the validation set contains 35, and
the test set contains 55 complexes. Each complex is a pair of ligand and receptor
proteins. A data sample is a pair of amino acids from different proteins and their interaction.
The total number of positive samples in the dataset is $20857$, which
is much less than the number of negative samples. The PN ratio is around 1:1000.
The Docking Benchmark version 4 (DB4)~\cite{hwang2010protein} and Docking Benchmark version 3 (DB3)~\cite{hwang2008protein} are two subsets of the DB5.
The DB4 contains 175 complexes and 16004 positive samples, and the DB3 contains 127 complexes and 12335 positive samples in total.

All dataset are mined from the Protein Data Back~\cite{10.1093/nar/28.1.235}.
Each protein has original sequential information from amino acid chains. The numbers of amino acid
in protein sequences ranges from tens to thousands.
The protein structures
are obtained from X-ray crystallography or biological mutagenesis experiments~\cite{afsar2014pairpred}.
In protein graphs, nodes are amino acid and edges are affinities between nodes.
both nodes and edges contain features from protein structures and sequences.

We use the same node features as in~\cite{fout2017protein,afsar2014pairpred}.
The node features are computed based on different properties of amino acid.
The residue depth is defined as the minimal distance for an amino acid
to the protein's surface. It's normalized in the range from 0 to 1
and has been demonstrated to carry valuable information for amino
acid interactions. The amino acid composition defines the count of a specific
amino acid in the direction and opposite direction of the side chain for the amino acid of interest.
The threshold along two directions is the minimal atomic distance of 8A. The
amino acid composition varies dramatically among amino acids, which
is vital to determine the properties of an amino acid. The protrusion index
for an amino acid is a collection of statistics of protrusion values for
all atoms along its side chain. These features deliver important inherent structural
information and properties for the amino acid of interest. They are combined and concatenated together in a consistent order, which results in the total number of node features to be 76.
We use the same edge features as in the work~\cite{fout2017protein}.
Each edge feature vector contains 2 features.
One is the normalized distance between two amino acids and
the other is the angle between two normal vectors for the two amino acid planes.

Recently a larger dataset Database of Interacting Protein Structures (DIPS)
is created by~\cite{townshend2019end}. An amino acid is represented as
4D grid-like data, which contains spatial information at the atom level and
types of all atoms in the amino acid. However, the structural information
is not considered in the dataset, thus the proteins can not be represented
as graphs.

\subsection{Baselines}
The baseline methods could be grouped into three categories, these are,
the state-of-the-art conventional machine learning method
BIPSPI~\cite{sanchez2019bipspi}, the
CNN-based method SASNet\cite{townshend2019end}
and the GNN-based methods DCNN~\cite{atwood2016diffusion}, NGF~\cite{duvenaud2015convolutional},
DTNN~\cite{schutt2017quantum} and NEA~\cite{fout2017protein}.
Particularly, the GNN-based baselines use different graph neural architectures
for node feature learning, but use the same dense layers as binary classifiers to predict
the interaction
for each pair of amino acid separately. \\
\textbf{BIPSPI} is the abbreviation for xgBoost Interface Prediction of
Specific-Partner Interactions. The method combines both structure
and sequence features and uses Extreme Gradient Boosting~\cite{chen2016xgboost} with a
novel scoring function for protein interface prediction.\\
\textbf{SASNet} is the Siamese Atomic Surfacelet Network, which uses only
spatial coordinates and types of all atoms in amino acids and voxelizes
all amino acids into a 4D-grid manner. The first three dimensions deliver
the spatial information of an amino acid and the last dimension is the one-hot
representation of types for all atoms in the amino acid. The paired two amino acid representations are then
passed to 3D CNN with weights sharing, followed by concatenation operation and
dense layers for binary classification to decide whether the two amino acids interact with each other.\\
\textbf{DCNNs} is diffusion-convolutional neural networks for graph-structured data applying
diffusion-convolution operators $k$ times ($k$-hops) for node feature learning.
A diffusion-convolution operator scans a diffusion process for each node.
For a node of interest,
$k$ diffusion-convolution operators gather information from all nodes that each of those can connect to the node of interest
through $k$ steps. Then several dense layers are used as a binary classifier to predict
the interactions of two nodes.\\
\textbf{NGF} is the commonly used graph convolutional networks, which first
aggregates node information in neighborhood by multiplying the adjacency matrix to the node feature matrix, and then
performs linear transformations on node features, followed by a nonlinear function for node feature learning.\\
\textbf{DTNN} is deep tensor neural networks, which aggregates both node and edge information
in neighborhood. Linear transformations are applied to node features and edge features separately.
After that, element-wise multiplication on the corresponding node features and edge features
is performed to achieve the final feature vector for the node of interest. Intuitively,
edges serve as gates to help control information from the corresponding nodes.\\
\textbf{NEA} is Node and Edge Average, the state-of-the-art
GNN-based method on the used DB5 dataset. Similar as \textbf{DTNN}, it performs aggregation
and linear transformation on both nodes and edge features in neighborhood. Then node and edge features
are averaged and summed together, followed by a residual connection and nonlinear
activation to generate node features.

\subsection{Experimental Setup}
We use the same data splitting as in all the baseline methods~\cite{fout2017protein,townshend2019end,sanchez2019bipspi} for the DB5 datasets.
For the DB4 and DB3 datasets, we first randomly split each dataset with a ratio of 6:2:2 for the training, validation and test samples,
then fix the splitting in all experiments.
We first only consider
the cross-protein pairwise interactions for training.
Similar to the
work~\cite{fout2017protein},  we keep all the positive examples and perform down-sampling on the negative samples, resulting in the PN ratio of 1:10 during the training phase. We maintain the original PN ratio in the validation and test phases.

For three GNN-based baselines, different numbers of GNN layers
are designed and explored in~\cite{fout2017protein}. We also conduct experiments using the same numbers of GNN layers for
fair comparisons. Our proposed GNN layer has two variants,
neighborhood average and neighborhood weighted average. We conduct experiments
on both for clear comparisons. For high-order pairwise interaction, we perform
concatenation on the feature vectors of two nodes from different proteins.
Several residual blocks are employed for dense prediction.
A residual block contains two 2D convolutional layers, the first of which is followed by ReLU
as the activation function. The number of intermediate channels of the first convolutional layer is
set as a hyperparameter. The identity map of the input is summed to the output of
the second convolutional layer, followed by ReLU to generate the final output.

We use the grid search to tune hyperparameters. The
search space for all hyperparameters is provided
in Table~\ref{tb:hyper}.
Adam Optimizer~\cite{kinga2015method} is employed for training
and ReLU is used as the activation function.
Each experimental setting is conducted to run 10 times
with different random seeds. All
hyperparameters are tuned based on the validation set.
Optimal
hyperparameters are tuned on one run and used across all
the 10 runs.

\begin{table}[t]
\begin{center}
\caption{
The search space for hyperparameters.
} \label{tb:hyper} {
\begin{tabular}{l|cccccccccccc}

\hline Hyperparameters &Search Space\\
\hline
\# of the Res. Blocks &3, 4, 5 \\
\# of Intermediate Channels &128, 192, 256\\
Learning Rate & e-1, 1e-2, 5e-3, 1e-3\\
Batch Size & 32, 64, 128 \\
\# of Epochs & 50, 80, 100 \\
Weight Decay & 1e-3, 1e-4, 1e-5 \\
Dropout & 0.3, 0.5, 0.8\\
\hline
\end{tabular}}
\end{center}
\end{table}

As positive examples and negative examples are not balanced,
we use Receiver operating characteristic (ROC) curve
for evaluation. Specifically, we calculate Area Under the ROC Curve (AUC)
based on the ROC curve for each complex. Then median AUC (MedAUC) for
all the complexes in the test sets is used to evaluate the performance of
different models. The used MedAUC can grantee very large or very small
proteins will not have dramatic effects on the performance on the whole dataset.

\subsection{Results}

\begin{table}[t]
\begin{center}
\caption{Comparison among different models in terms of MedAUC.
All the GNN-based methods apply one GNN layer for fair and convenient comparison.
For the DB5 dataset, results for
all the baselines are directly reported from the papers~\cite{fout2017protein,townshend2019end}.
The best performance is in bold.
} \label{tb:main_results} {
\begin{tabular}{l|cccccccccccc}

\hline Method &DB5 &DB4 &DB3\\
\hline
BIPSPI &0.878 (0.003) &0.882 (0.004) &0.891 (0.016) \\
SASNet & 0.876 (0.037) & 0.866 (0.025) & 0.862 (0.011)\\
DCNN & 0.828 (0.018) & 0.843 (0.022)& 0.858 (0.015)\\
NGF & 0.865 (0.007) & 0.879 (0.017)& 0.867 (0.016) \\
DTNN & 0.867 (0.007) & 0.868 (0.013) & 0.883 (0.008)  \\
NEA & 0.876 (0.005) & 0.884 (0.009) & 0.881 (0.014)\\
NeiA+HOPI &  \textbf{0.902 (0.012)} & \textbf{0.916 (0.014)} & \textbf{0.910 (0.009)}\\
NeiWA+HOPI & \textbf {0.908 (0.019)}& \textbf{0.921 (0.018)} & \textbf{0.913 (0.013)})\\
\hline
\end{tabular}}
\end{center}
\end{table}

\begin{table*}[t]
\begin{center}
\caption{Comparison among the GNN-based methods in terms of MedAUC on the DB5 dataset.
The number of GNN layers varies from 1 to 4. For all baseline methods,
We report the results taken from the paper~\cite{fout2017protein}.
The best performance is in bold.
} \label{tb:diff_gnns} {
\begin{tabular}{l|ccccc|ccccccc}
\hline \multirow{2}{*}{Method}  & \multicolumn{4}{c}{Number of GNN Layers} \\
& 1 & 2 & 3 & 4  \\
\hline
NGF &  0.865 (0.007) &  0.871 (0.013) & 0.873 (0.017) & 0.869 (0.017) \\
DTNN & 0.867 (0.007) & 0.880 (0.007) & 0.882 (0.008) & 0.873 (0.012)\\
Node and Edge Average & 0.876 (0.005) & 0.898 (0.005) & 0.895 (0.006) & 0.889 (0.007) \\
NeiA+HOPI &  \textbf{0.902 (0.012)} & \textbf{0.919 (0.015)} & \textbf{0.921 (0.009)} & \textbf{0.915 (0.009)}  \\
NeiWA+HOPI &  \textbf{0.908 (0.019)} & \textbf{0.930 (0.016)} & \textbf{0.924 (0.011)} & \textbf{0.914 (0.013)} \\
\hline
\end{tabular}}
\end{center}
\end{table*}

\begin{table}[b]
\begin{center}
\caption{
Performance of incorporating in-protein pairwise
interactions on the DB5 dataset. The original NeiA+HOPI and NeiWA+HOPI methods
without in-protein pairwise interactions serve as baselines.
`w/o' denotes `without' and `w' denotes `with'.
} \label{tb:in_protein} {
\begin{tabular}{l|cccccccccccc}

\hline Method &Ratio&MedAUC\\
\hline
NeiA+HOPI w/o in-protein & 1:10 &0.902 (0.012) \\
\multirow{4}{*}{NeiA+HOPI w in-protein} & 1:7 & 0.911 (0.017)\\
& 1:5 & 0.910 (0.017)\\
& 1:3 & 0.901 (0.014)\\
& 1:1 & 0.896 (0.013)\\
\hline NeiWA+HOPI w/o in-protein& 1:10 &0.908 (0.019) \\
\multirow{4}{*}{NeiWA+HOPI w in-protein} & 1:7 & 0.915 (0.021)\\
& 1:5 & 0.913 (0.017)\\
& 1:3 & 0.910 (0.018)\\
& 1:1 & 0.898 (0.013)\\
\hline
\end{tabular}}
\end{center}
\end{table}

\subsubsection{Performance Study}
We apply both variants of our GNN layers with the same sequential
modeling and high-order pairwise interaction methods, denoted as NeiA+HOPI
and NeiWA+HOPI, respectively. In
this section, we compare our approaches with
several baselines in terms of MedAUC on the three datasets.
We fix the number of GNN layers for all GNN-based methods to be 1 for
convenient comparisons, and
the results are reported in Table~\ref{tb:main_results}.
Note that for all baseline approaches on the DB5 dataset, we report the results
taken from papers~\cite{fout2017protein,townshend2019end}.
As the results for DB4 and DB3 are not reported in the related papers, we use
the same data splitting and run experiments for all methods.
All experiments run 10 times with random seeds. The average and the
standard deviation of testing MedAUCs across the 10 runs are reported.

We can observe that our proposed approaches outperform all
the baselines significantly. Specifically, the performance of our
NeiA+HOPI is $2.6\%$, $3.2\%$, $2.9\%$
higher than the previous best GNN-based method NEA method over three datasets.
Surprisingly, our proposed NeiWA+HOPI outperforms NEA by a larger margin
of $3.2\%$, $3.7\%$ and $3.2\%$ on the three datasets, respectively.
The NeiA+HOPI and NeiWA+HOPI also exhibit considerable improvement compared with the conventional machine learning method BIPSPI and
CNN-based method SASNet.
Note that the main difference of the NeiA+HOPI compared with NEA
is the use of our proposed SM and HOPI methods.
We preserve the original sequential information in proteins
and use CNNs to capture the high-level pairwise interaction patterns. The superior
performance of the NeiA+HOPI demonstrates the effectiveness
of our proposed SM and HOPI methods. Different from the GNN-based
methods, SASNet uses 3D convolution layers for feature extraction
and then applies dense layers for binary classification.
It leverages 3D spatial information of amino acids at the atom level but ignores the structural information. Our methods
explicitly consider the structural and sequential information and high-order pairwise interactions,
thereby leading to much better performance for protein interface prediction.

The four GNN-based methods use
the same dense layers for binary classification but differ in
graph neural architectures. Compared with NGF and DTNN, NEA
incorporates additional edge information from neighborhood. DTNN performs element-wise multiplication
but NEA performs summation
over a node feature matrix and the corresponding edge feature matrix. Our methods make use of the information from edges by adding
it to node features for powerful node representations. Basically, NEA computes
the feature vector for the node of interest by averaging nodes and edges from its
neighborhood. The assumption here is that all nodes and edges contribute equally to the center node.
The NeiWA+HOPI selects more important nodes and edges by assigning larger weights to them,
resulting in a slight improvement in performance compared with the NeiA+HOPI.

\subsubsection{Comparison with GNN-based Methods}
One GNN layer can incorporate 1-hop information from
neighborhood to node features. Stacking $k$ GCN layers is capable of enlarging
receptive fields by aggregating $k$-hops information.
It's suggested that applying several GCNs layers can improve the interface prediction
for some graph neural architectures~\cite{fout2017protein}. To explore such properties in
our models and provide fair comparisons, we apply different numbers of GNN layers and conduct experiments on the DB5 dataset.
The results are reported in Table~\ref{tb:diff_gnns}.
We can observe from the table that our methods achieve the best performance
despite the number of GNN layers.
This again demonstrates the effectiveness of
our proposed SM and HOPI methods.
Note that the other three GNN-based methods give better results when
the number of GNN layers increases to 2 and 3, but start to harm the
performance when it reaches 4. Consistent observations are shown in our models.
Apparently, the model capacity of graph neural architectures can reach the upper bound
but the proposed SM and HOPI help extract the sequential information and explore the inherent high-order
pairwise interactions for accurate interface prediction.

\subsubsection{Affect of In-Protein Pairwise Interactions}
As the number of positive examples is relatively small in cross-protein amino acid
pairs, we conduct experiments on the DB5 dataset and add some positive in-protein pairs in the training process.
We keep the number of positive cross-protein pairs unchanged.
For each complex, we randomly select the same number of positive examples in the ligand protein
and the receptor protein.
The final PN ration is set to be 1:7, 1:5, 1:3 and 1:1, respectively.
The experimental results are shown in Table~\ref{tb:in_protein}.
We can observe from the results that the performance increases
when adding positive in-protein examples and making the PN
ratios to be 1:7 and 1:5. When more positive in-protein examples
are added for training and the PN ratio reaches 1:1, the performance starts
to decrease and becomes worse than that without in-protein pairs.
This indicates that the inherent properties of amino acids may affect
the interactions between them. These in-protein interactions are beneficial
to the prediction of cross-protein interactions.
However, when the in-protein interactions become dominant through adding too much
positive in-protein examples, the prediction of cross-protein interactions is somehow
interfered and harmed.




\section{Conclusion}
We study protein interface prediction.
The latest state-of-the-art method represents
proteins as graphs, but fails to consider sequential information from amino acid chains.
We propose a novel model to incorporate both structural and sequential information,
and high-order pairwise interactions for accurate interface prediction.
We generate a 3D tensor to store these information. The output is adapted to a 2D map containing
interactions for all amino acid pairs. The task becomes
a 2D dense prediction task, where 2D convolutional neural
networks are employed to learn high-level interaction patterns.
We evaluate our methods over different datasets. The experimental results
demonstrate the effectiveness of our proposed approach.

\begin{acks}
This work was supported in part by National Science Foundation grants IIS-1908220 and DBI-1922969.
\end{acks}

\end{document}